\documentclass{article}
\usepackage{spconf,amsmath,graphicx}
\usepackage[table]{xcolor}
\usepackage{multirow}
\usepackage{tkz-tab}
\usepackage{cleveref}
\usepackage{booktabs}
\usepackage{titlesec}
\usepackage{pdfpages} 
\usepackage{subcaption}
\usepackage{caption}
\usepackage[style=apa, backend=biber]{biblatex}  
\newcommand{\myheading}[1]{%
  \noindent{\normalsize\bfseries #1}\\[0.5em]%
  \noindent
}

\DeclareMathOperator{\STI}{\mathit{STI}}  
\DeclareMathOperator{\SNR}{\mathit{SNR}}  

\title{Room acoustics affect communicative success in hybrid meeting spaces: a pilot study}
\name{
    Robert Einig\textsuperscript{1}, 
    Stefan Janscha\textsuperscript{1}, 
    Jonas Schuster\textsuperscript{1}, 
    Julian Koch, 
    Martin Hagm\"uller, 
    Barbara Schuppler\textsuperscript{2} \thanks{$^1$\ These authors contributed equally to this work} \thanks{$^2$\ Corresponding author}
    }
    
\address{
    Signal Processing and Speech Communication Laboratory, Graz University of Technology \\
    \small{\{einig, stefan.janschaivan, jonas.schuster\}@student.tugraz.at} \\
    \small{\{julian.koch, hagmueller, b.schuppler\}@tugraz.at}
    }

\begin{document}
\maketitle
\thispagestyle{plain}   
\pagestyle{plain}       

%


\begin{abstract}
Since the COVID-19 pandemic in 2020, universities and companies have increasingly integrated hybrid features into their meeting spaces, or even created dedicated rooms for this purpose. While the importance of a fast and stable internet connection is often prioritized, the acoustic design of seminar rooms is frequently overlooked. Poor acoustics, particularly excessive reverberation, can lead to issues such as misunderstandings, reduced speech intelligibility or cognitive and vocal fatigue. This pilot study investigates whether room acoustic interventions in a seminar room at  Graz University of Technology support better communication in hybrid meetings. For this purpose, we recorded two groups of persons twice, once before and once after improving the acoustics of the room. Our findings -- despite not reaching statistical significance due to the small sample size - indicate clearly that our spatial interventions improve communicative success in hybrid meetings. To make the paper accessible also for readers from the speech communication community, we explain room acoustics background, relevant for the interpretation of our results.
\end{abstract}


\section{Introduction}
\label{sec:intro}
In recent years, the use of video conferencing tools has become increasingly popular. While Skype dominated the market for a long time, newer applications such as Zoom, Microsoft Teams and webex have gained widespread use. Nowadays, it is common for universities and companies to include such tools in their daily work environment. Despite the ease of setting up these tools, the quality of communication tends to suffer. Insufficient audio and video quality often result in a higher cognitive effort and mental fatigue \parencite{fatigue}. These issues often lead to a less productive, attentive and collaborative engagement between the attendees. 
\par
How can we support communication in hybrid settings from a technical point of view? While video quality is something you can improve by replacing the equipment, audio quality is more difficult to address, as it involves complex challenges. A simple replacement of your microphone with a better one is often insufficient and audio quality remains unsatisfactory. However, the issues often do not arise from the equipment itself, but from poor room acoustics and a bad spatial setup \parencite{review, fels}. Addressing acoustical issues is crucial for a productive meeting environment, enabling participants to communicate, concentrate and work efficiently. 
\par
The aim of this study is to compare communicative success during meetings that involve a video conferencing tool, with a particular focus on the acoustics of the seminar room and the impact of acoustical modifications. To evaluate communicative success, we use both qualitative data from participant questionnaires and standard measures for assessing the room acoustics. Specifically, this study addresses two research questions: (RQ1) To what extent does the room, in its current state, meet the requirements for effective use in a hybrid meeting? (RQ2) Does communication success in hybrid meetings improve when room acoustics are enhanced? To answer these questions, we perform room acoustic measurements and design interventions to improve the acoustics. Additionally, we conduct experiments with participants in both treated and untreated rooms, in which they perform a communicative task. During these experiments, two participants are physically present in the room, while one participates remotely. Combining room acoustic measurements with questionnaires on perceived communicative quality allows us to assess whether the acoustic treatments affect communication quality in hybrid meetings.

\subsection{Improved Room Acoustics Quality in Meeting Rooms}
To achieve suitable meeting room conditions for speech communication, adequate acoustic treatment and noise prevention are essential. Objective evaluation typically involves measuring reverberation time ($T_{30}$), speech clarity ($C_{50}$), and the speech transmission index ($\STI$), all of which can be derived from room impulse responses. These parameters provide a quantitative basis for assessing whether a space supports intelligible communication or imposes unnecessary cognitive strain on participants.
\par
Several studies have established the relevance of these measures in educational and professional contexts. For instance, \textcite{bachelor} demonstrated their application in classrooms, while reviews highlight their widespread use in building and room design to improve speech communication when participants share the same physical space \parencite{review, fels}. \textcite{sti} further provided methodological detail and reference values, proposing $C_{50}$ values above 0,dB and high $\STI$ scores as indicators of good acoustic quality. 
\par
Applied research has demonstrated that targeted interventions can substantially improve acoustic measures. \textcite{Improved} examined the effects of acoustic treatments in meeting rooms with volumes between $300 - 500,\text{m}^3$. Their findings show that applying sound-absorbing materials reduced reverberation times by approximately $40\%$, thereby enhancing speech clarity, while the addition of diffusion panels improved the spatial distribution of sound. As a general guideline for absorber and diffuser placement, the authors recommend covering $50 - 70\%$ of the ceiling and $20 - 40\%$ of the walls, including the rear wall, with sound-absorbing material in regularly shaped rooms of about $500,\text{m}^3$. For smaller meeting rooms of around $300,\text{m}^3$, $20 - 30\%$ of the ceiling and up to $15\%$ of the rear walls should be treated \parencite[p. 471]{Improved}. Following such treatments, users reported improved perceived audio quality and a reduction in cognitive load.
\par
Findings from the educational sector further emphasize the importance of acoustic design. \textcite{academic} conducted a large-scale study of 45 unoccupied university classrooms in the Netherlands (15 lecture halls, 16 regular classrooms, 14 skills laboratories) to assess compliance with recommended reference levels for speech intelligibility. Measurements of background noise and the Speech Transmission Index ($\STI$) showed that 41 classrooms exceeded the maximum noise level of $35,\text{dB(A)}$, with 17 surpassing $40,\text{dB(A)}$. Only six classrooms achieved “excellent” intelligibility ($STI\geq 0.75$) at a five-meter distance, and none reached this level at more representative, disadvantageous listening positions. At these positions, the mean $\STI$ was $0.61$, and 24 classrooms fell below the minimum recommended threshold of $0.61$. The study concluded that most classrooms provide insufficient speech intelligibility, highlighting the need for $\STI$ values of at least $0.75$ throughout the room in higher education, where complex academic language and non-native instruction increase cognitive demands.

\subsection{Speaking and listening in video conferences}
Cognitive overload and mental fatigue represent major challenges in videoconferencing. In a study with 35 participants, \textcite{fatigue} compared a 50-minute engineering lecture delivered face-to-face and via video conference. Throughout the lectures, neurophysiological data were collected, including Event-Related Potentials (ERP), Heart Rate Variability (HRV), and fatigue indicators based on EEG and ECG. Participants also provided self-reports after each session. The results showed that videoconferencing led to significantly higher levels of fatigue, drowsiness, and tiredness compared to face-to-face lectures, with mood deterioration also frequently reported. Neurophysiological measurements confirmed these findings, revealing elevated HRV and reduced attentional resources as reflected in ERP changes during video conferences. Based on these outcomes, the authors recommend that videoconferencing be used as a complement rather than a replacement for in-person interactions, given its fatigue-inducing effects.
\par
From a theoretical perspective, the Ease of Language Understanding (ELU) model \parencite{elu} provides a relevant framework for investigating listening fatigue. The model emphasizes the crucial role of working memory (WM) in language processing, particularly in conversational and noisy contexts. WM enables listeners to retain relevant information, filter out distractions, and focus on key aspects of a discussion. According to ELU, when speech input closely matches stored phonological representations, comprehension occurs automatically with little effort. However, in adverse listening conditions—such as background noise, hearing loss, or misunderstandings—this automatic process suffers. Listeners must then rely on controlled WM processes to resolve mismatches, which require additional cognitive resources. Because working memory capacity (WMC) is limited, these compensatory processes increase listening effort and contribute to mental fatigue, especially when speech is complex or key elements of a conversation are difficult to access.
\par
Communication in videoconferencing settings can be challenging not only for listeners but also for speakers, as it may affect the effectiveness and perceived charisma of their voice and speech. \textcite{Charisma} investigated the negative impact of audio compression on vocal quality, comparing differences in degradation between male and female voices. Anonymous surveys using various audio formats were conducted to assess perceived vocal charisma. The findings were clear: while the charisma of male voices decreased by an average of $6.5\,\%$, female voices experienced a substantially greater reduction of $20\,\%$. These results highlight the importance of including both male and female voices in testing to ensure improvements are broadly applicable.
\clearpage
\newpage

\section{Improving Room Acoustics for Hybrid Meetings}

\subsection{Methodology}
Our first step was to perform an acoustic measurement of room IDEG134, located at Inffeldgasse 16c, Graz University of Technology. This space is primarily used as a lecture room and a hybrid seminar room, as it is equipped with an integrated Webex Board. It serves as the experimental setting for all investigations in this study. Additionally, the room contains a projector, which produces most of the ambient noise. We assessed the room's acoustic properties by measuring the impulse response at different seating locations and the noise level generated by the projector.

\subsubsection{Geometric Measurements}
Prior to commencing the acoustic measurement procedure, we determined the seminar room’s volume and compiled a comprehensive inventory of its furnishings. The measured volume of the room is $150.8\,\text{m}^3$. The room deviates from a standard rectangular layout. Figures \ref{fig:floorplan-Top} and \ref{fig:floorplan-Side} show the scaled floor plan in top-down and front views, respectively. The seminar room features a blackboard at the front, and two large monitors with a speaker are located at the back, where Webex meetings are held. Figure \ref{fig:Webex_treated} shows the Webex setup at the back of the room. An object list identified 29 chairs and 13 tables within the space.

To establish the valid frequency range for subsequent acoustic data analysis, we calculated the Schroeder frequency. This frequency is derived from the room volume and an estimated reverberation time. For the seminar room under study, we estimated a reverberation time of $0.6,\text{s}$ and thereby determined a Schroeder frequency of $126.2\,\text{Hz}$. Below this frequency, the sound field in the room is dominated by room modes, meaning that sound levels are not evenly distributed, and acoustic measurements in this range must be interpreted with caution \parencite[]{oe_8115}.

\subsubsection{Room Acoustic Measurements}
We began by characterizing the room acoustics through impulse response measurements in two scenarios: untreated and acoustically treated. The untreated measurements served as a baseline for the room’s natural acoustic properties and guided the selection of the most effective treatment strategy. Following the installation of acoustic treatment, measurements were repeated to quantify improvements. From these data, we evaluated the reverberation time ($T_{30}$), speech clarity index ($C_{50}$), and Speech Transmission Index ($\STI$), which are essential parameters for objectively assessing speech intelligibility. Impulse response measurements and parameter calculations were performed using the ITA-Toolbox in MATLAB \parencite{ita-toolbox}. Given the significant impact of projector noise on speech intelligibility, we additionally measured the sound pressure level, a critical factor for accurate $\STI$ calculations.
\par
We placed our sound source in typical speaker locations: at the Webex Board (S1), at the blackboard (S2), and in the center of the room (S3). Impulse response measurements followed the procedure outlined in  \textcite{oe_3382-1}, but we intentionally relaxed some of the strict requirements to better reflect realistic scenarios. 

\begin{minipage}[t]{0.45\textwidth}
    \centering
    \includegraphics[width=0.95\linewidth, trim=0 10 0 10,clip]{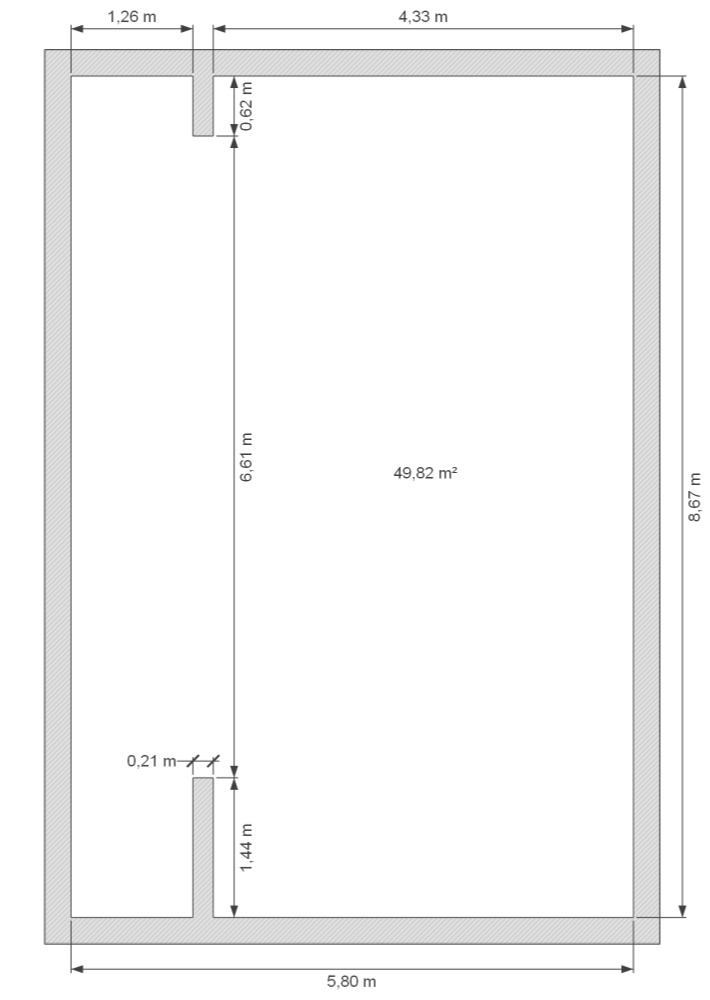}
    \captionof{figure}{Scaled floor plan of the seminar room in top-down view, showing wall dimensions and overall room geometry.}
    \label{fig:floorplan-Top}
\end{minipage}

\begin{minipage}[t]{0.45\textwidth}
    \centering
    \includegraphics[width=0.95\linewidth, trim=0 0 0 10,clip]{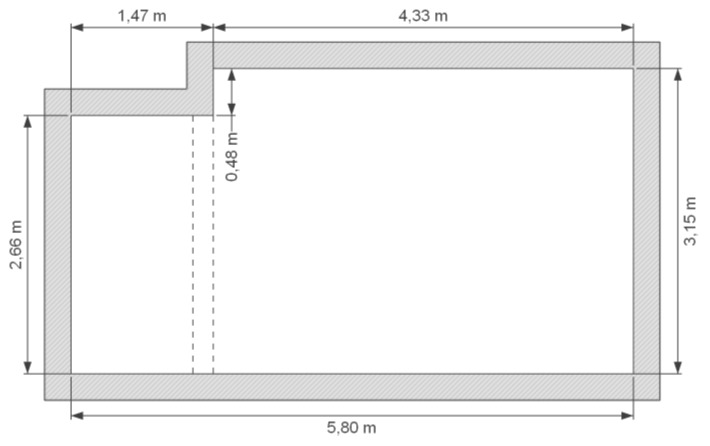}
    \captionof{figure}{Scaled floor plan of the seminar room in front view, showing the height dimensions of the room.}
    \label{fig:floorplan-Side}
\end{minipage}

\begin{figure*}[t]
    \centering
    \includegraphics[width=\linewidth]{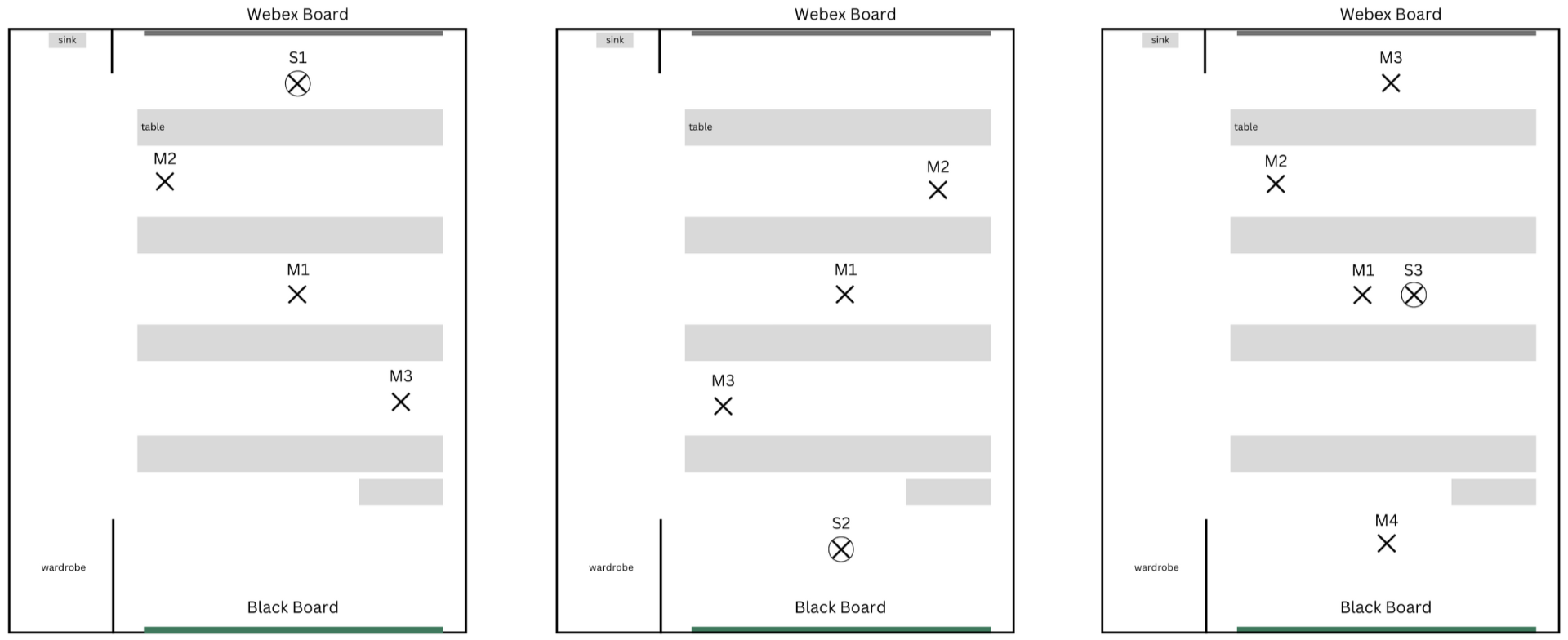}
    \captionof{figure}{Schematic of the three individual acoustic measurement setups displaying our sound source (S1, S2, and S3) and receiver (M1, M2, M3, and M4) placements in the seminar room. The three measurement setups are independent from each other.}
    \label{fig:Meas_setup}
\end{figure*}

For instance, when placing the sound source at the Webex Board, the loudspeaker was positioned close to the wall, and when it was in the center of the room, one microphone was placed right next to it to simulate a listener sitting nearby. The source at the blackboard was positioned according to the standard while still maintaining realism. For each condition, we set three to four distinct receiver positions to capture the impulse response. Measurements were performed using an exponential sine sweep lasting 10\,s. Figure \ref{fig:Meas_setup} illustrates our setup, with S1, S2, and S3 indicating loudspeaker positions and M1, M2, M3, and M4 showing microphone placements.

We used a Nor276 Dodecahedron loudspeaker as our sound source and a single NTi M32 measurement microphone as the receiver. Since only one microphone was available, each receiver position had to be measured separately. Both devices were connected to a Focusrite 2i2 2nd Gen audio interface. 
\par
After completing the impulse response measurements, we measured the projector’s sound level at each receiver position using an NTi Audio XL2 sound level meter. Measurements were conducted in octave bands and time-weighted over $15\,\text{s}$.
\par
\subsubsection{Room Acoustic Intervention}
\begin{figure*}[!htb]
   \centering
   \includegraphics[width=0.95\linewidth]{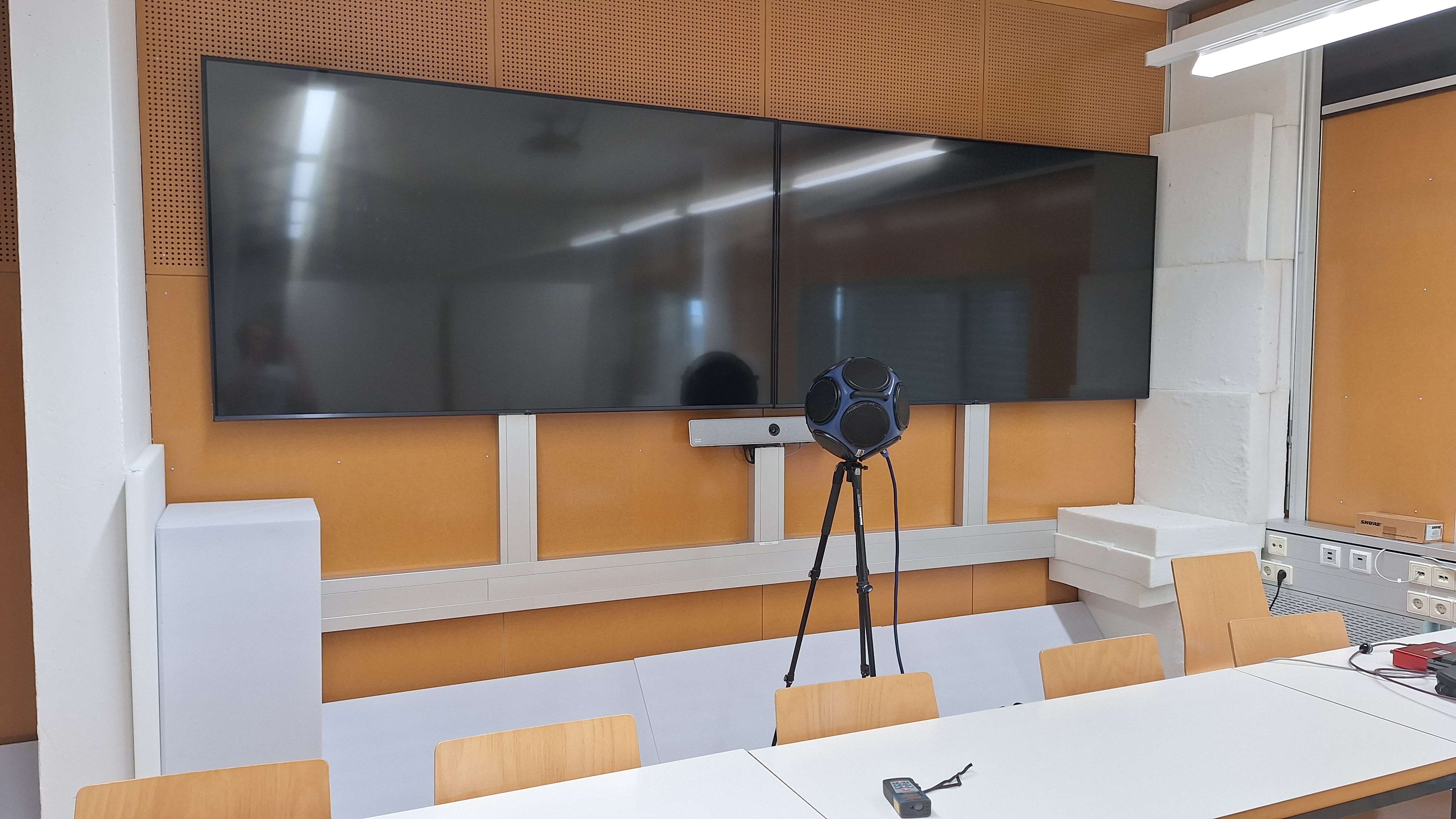}
   \captionof{figure}{Acoustic measurement after room acoustic intervention with sound source (S1) positioned at Webex board. Bass traps and absorbers are stacked on each other and placed close to room edges.}
   \label{fig:Webex_treated}
\end{figure*}

\begin{figure*}[!htb]
   \centering
   \includegraphics[width=0.95\linewidth]{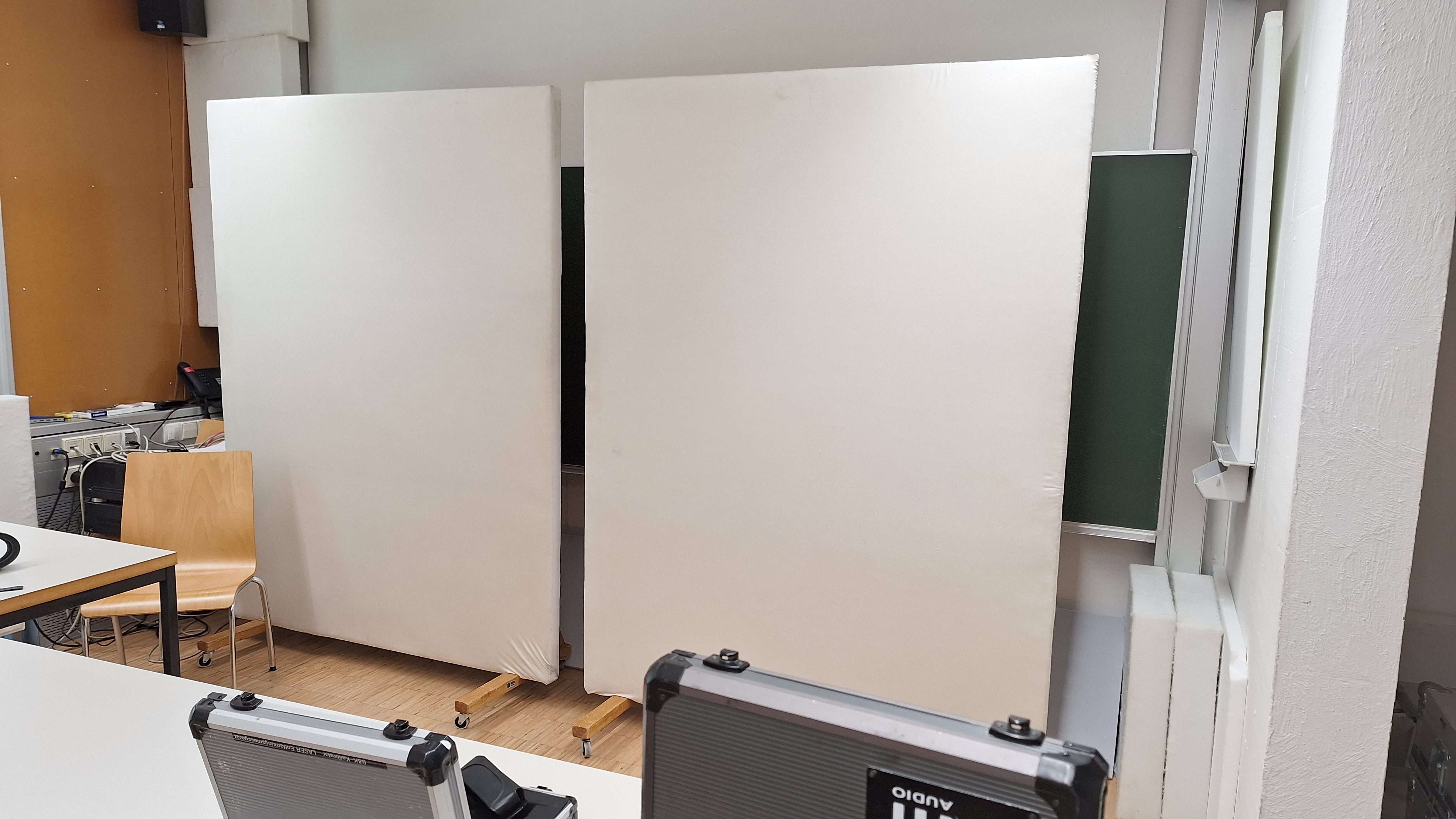}
   \captionof{figure}{Acoustic panels placed in front of the blackboard for measurement conditions S1 (source at Webex Board) and S3 (source in center of room). For the S2 measurement (source positioned at blackboard) we moved the panels to the right hand side of the room from the picture's point of view.}
   \label{fig:Panels}
\end{figure*}

To improve the room's acoustic qualities, we installed suitable acoustic absorbers. These absorbers were used as test absorbers and were not permanently mounted to the walls. Based on our initial measurement results, we found that reverberation times below $200\,\text{Hz}$ were excessively high across all measurement positions. This indicates a significant low-frequency reverberation issue that requires acoustic treatment. The peak in $T_{30}$ below $200\,\text{Hz}$, combined with the Schroeder frequency of $126.2\,\text{Hz}$, indicates that wave phenomena (like room modes and interference) dominate the room's acoustic behavior in this range \parencite[]{edge-absorbers}. Due to this finding, we targeted placing absorbers close to edges in the room. Placing porous absorbers in these locations effectively dampens room modes because acoustic pressure maxima occur there \parencite[]{edge-absorbers}. Specifically, we used multiple porous bass traps (14x) and surface absorbers (39x). Figure \ref{fig:Webex_treated} shows some of these absorbers stacked close to room edges in the back. Furthermore, we used two acoustic panels to dampen mid to high frequencies. We placed the acoustic panels in front of the blackboard for the two measurements, with the sound source each at the Webex Board and in the center of the room. For the other measurement position with the sound source placed at the blackboard, we moved them to the side of the room. Figure \ref{fig:Panels} shows our acoustic panels in front of the blackboard.

\subsubsection{Room Acoustic Parameter Calculation}
For evaluating our room acoustic measurements, we calculated $T_{30}$, $C_{50}$ and $\STI$. These parameters are crucial for objectively rating a room in terms of speech intelligibility. 

\paragraph{Reverberation Time $T_{30}$} The reverberation time quantifies how long it takes for sound to decay in an enclosed space. Specifically, $T_{30}$ is calculated by measuring the time it takes for the sound decay curve to fall from $5\,\text{dB}$ to $35\,\text{dB}$ below initial sound level, and then extrapolating this $30\,\text{dB}$ decay to a $60\,\text{dB}$ decay \parencite[]{oe_3382-1}. Optimal reverberation values, such as those recommended by standards like \textcite[]{oe_8115}, are defined based on the room's volume and its intended use. The mentioned standard provides target values and acceptable tolerance ranges for reverberation time across different frequency bands. 

\paragraph{Clarity Index $C_{50}$} The Clarity Index is an acoustic parameter used to assess speech intelligibility and sound quality in enclosed spaces. It is defined as the logarithmic ratio (in dB) of early arriving energy (within the first $50\,\text{ms}$ after the direct sound) to the later arriving energy (after $50\,\text{ms}$ to infinity) \parencite[]{Improved}. A high $C_{50}$ value indicates that most of the speech signal arrives early, enhancing intelligibility. A low value suggests excessive reverberation, making speech more difficult to understand. We calculated our $T_{30}$ and $C_{50}$ values in MATLAB with the ITA-Toolbox \parencite[]{ita-toolbox} via our measured impulse responses. After that, we averaged all results per measurement condition to receive the spatial average.

\paragraph{Speech Transmission Index ($\STI$)} 
The $\STI$ is an objective measure designed to predict speech intelligibility from a sound source to a listener. It quantifies how noise, reverberation, and signal distortions affect the clarity of speech by analyzing the modulation transfer function ($MTF$) across different frequency bands. $\STI$ operates on a scale from 0 (not intelligible) to 1 (perfectly intelligible speech). Common classifications range from "bad" ($\leq 0.30$) to "excellent" ($\geq 0.75$) for normal-hearing adults in daily conversation \parencite[]{IEC60286-16}. Table \ref{tab:STI-ratings} shows the classification of $\STI$ values. 
\par
The calculation of $\STI$ considers seven speech-important frequency bands (from $125\,\text{Hz}$ to $8000\,\text{Hz}$) and 14 modulation frequencies (between $0.63\,\text{Hz}$ and $12.5\,\text{Hz}$) within each band. This process determines the modulation reduction factor by comparing the modulation depth of a test signal at the input to its reduced modulation at the output. This is directly related to the effective signal-to-noise ratio ($\SNR$) \parencite[]{sti}. The $\STI$ method also incorporates octave-band weighting factors ($\alpha$) and redundancy factors ($\beta$), which are gender-specific for male and female speech to improve prediction accuracy. 
\par
We calculated the $\STI$ according to \cite{IEC60286-16}. For our calculation method, we chose to compute the $\STI$ indirectly via our measured impulse responses. This method requires a noise-free impulse response and the ambient noise level of the seminar room. Because the $\STI$ is heavily influenced by the $\SNR$, it is important to consider the noise generated from the projector. Due to the noise level generated by the projector being substantially louder than the general background noise level, we neglected the latter and used the measured projector noise as the relevant ambient noise input. We conducted the impulse response and noise level measurements separately. Therefore, we assumed that our impulse response measurements are noise-free.
\par
Since we assumed that our impulse responses are noise-free, we were able to use the average speech level of normal vocal effort outlined in ISO 3382-3 for our calculation process of the modulation transfer function $MTF$ \parencite{oe_3382-3}. Furthermore, we corrected the speech level of each microphone position to account for the distance-dependent attenuation of the sound source.

\begin{table}
    \captionsetup{position=above}
    \caption{STI intelligibility ratings based on IEC 60286-16:2021 \parencite[]{IEC60286-16}.}
    \centering
    \begin{tabular}{cc}\toprule
         Intelligibility Rating& STI\\\midrule
         Bad& $0.00 - 0.30$\\
         Poor& $0.30 - 0.45$\\
         Fair& $0.45 - 0.60$\\
         Good& $0.60 - 0.75$\\
         Excellent& $0.75 - 1.00$\\ \bottomrule
    \end{tabular}
    \label{tab:STI-ratings}
\end{table}

\subsection{Room Acoustic Measurement Results}

\subsubsection{Reverberation Time $T_{30}$}
To evaluate the $T_{30}$ results of our acoustic measurement, we referred to \textcite{oe_8115}. Although we did not strictly follow the procedure of this standard, it still provides a useful guideline for assessing our results. $T_{30}$ requirements heavily depend on room usage. Our room is mainly used as a hybrid meeting room ("Kommunikation Klasse C") and as a lecture classroom ("Sprachdarbietung Klasse B"). Both of these categories recommend a reverberation time of $0.5\,\text{s}$ with a $\pm\,20\,\%$ tolerance for a room volume of $150.8\,\text{m}^{3}$ in terms of speech intelligibility. For lower frequencies, a higher tolerance range is tolerable. 
\par
\begin{figure*}[t]
    \centering
    \begin{subfigure}[b]{0.45\textwidth}
        \includegraphics[width=\textwidth]{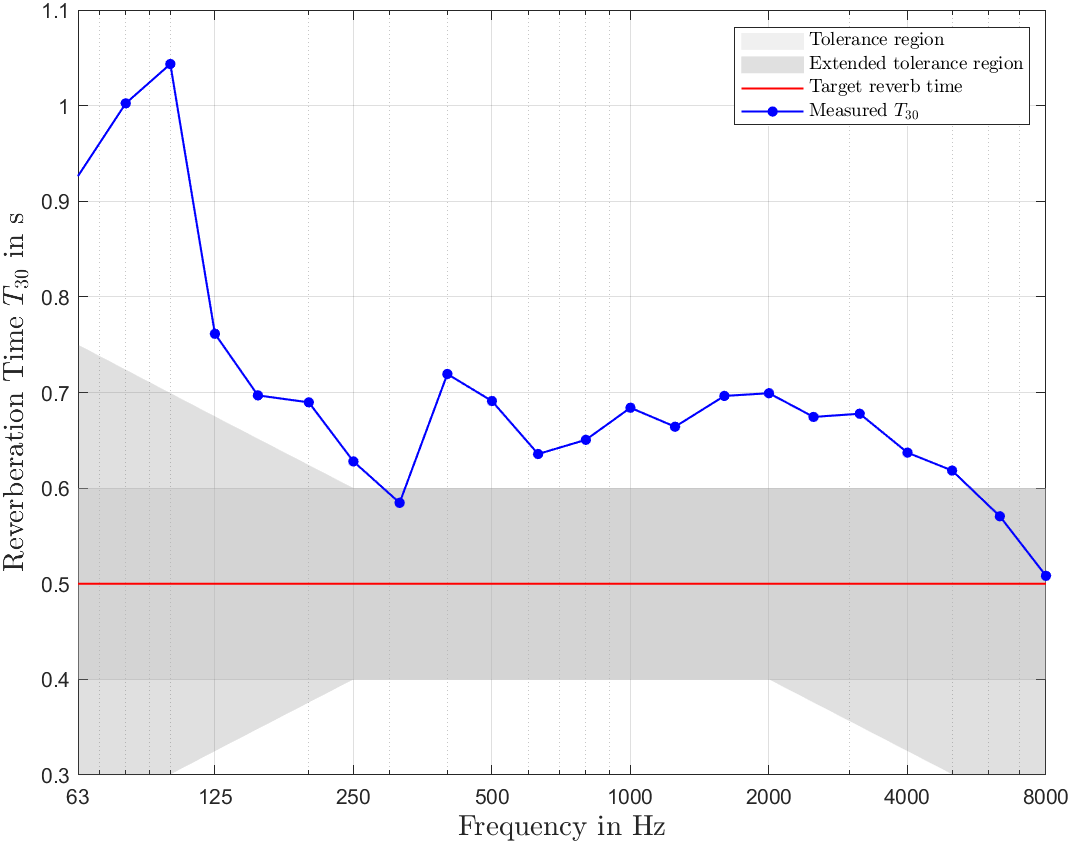}
        \caption{S1 (Webex): Before room acoustic intervention.}
        \label{fig:T30_a}
   \end{subfigure}
   \hfill
    \begin{subfigure}[b]{0.45\textwidth}
        \includegraphics[width=\textwidth]{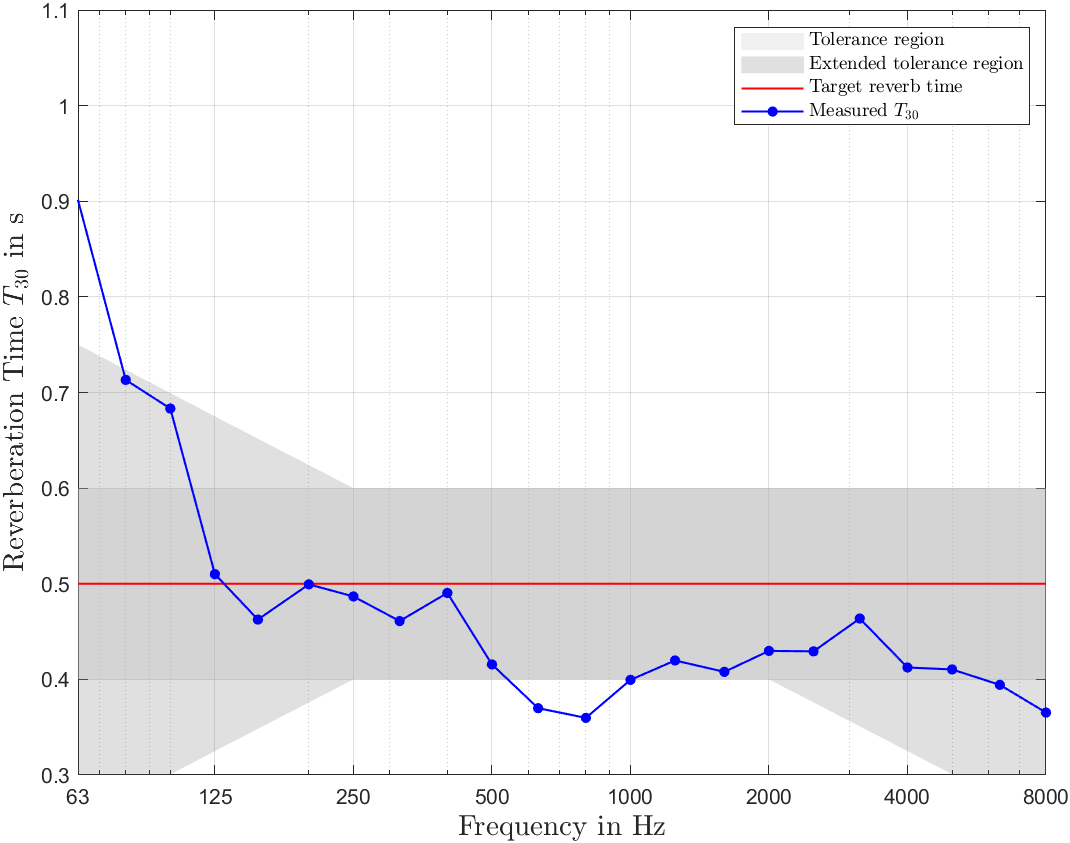}
        \caption{S1 (Webex): After room acoustic intervention.}
        \label{fig:T30_b}
   \end{subfigure}

   \centering
   \begin{subfigure}[b]{0.45\textwidth}
       \includegraphics[width=\textwidth]{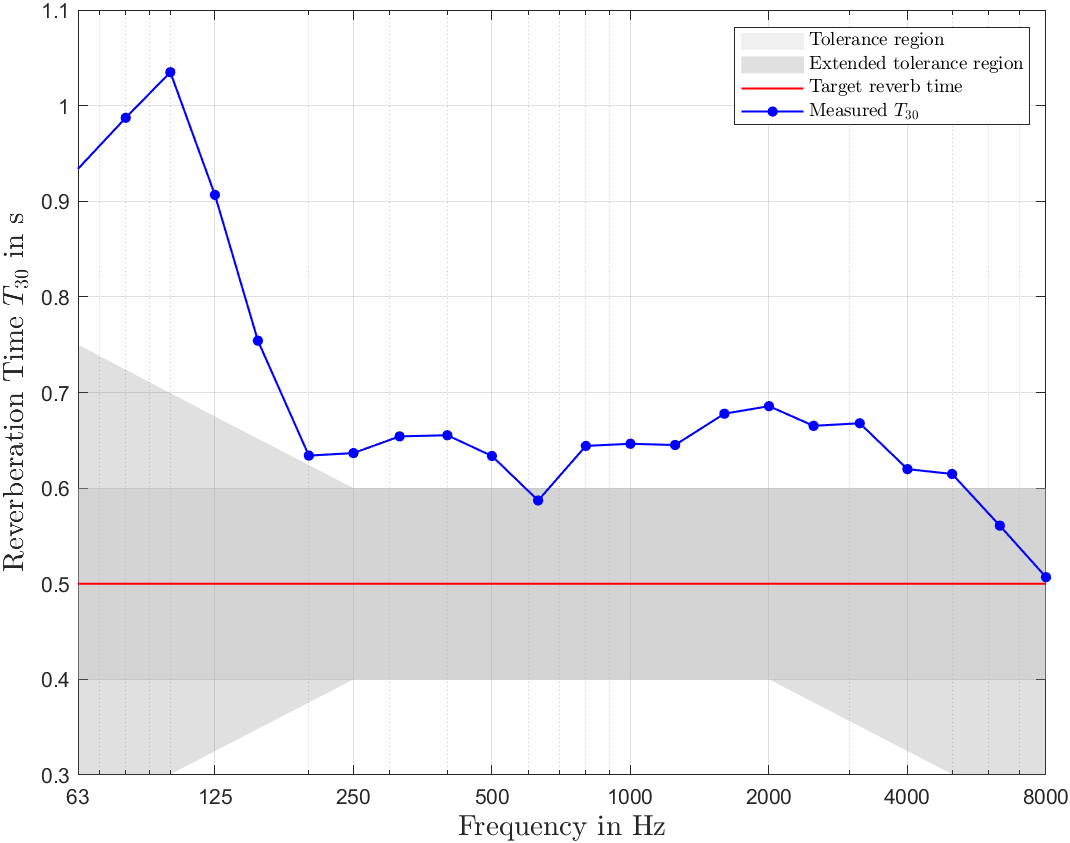}
       \caption{S2 (Blackboard): Before room acoustic intervention.}
       \label{fig:T30_c}
   \end{subfigure}
   \hfill
   \begin{subfigure}[b]{0.45\textwidth}
       \includegraphics[width=\textwidth]{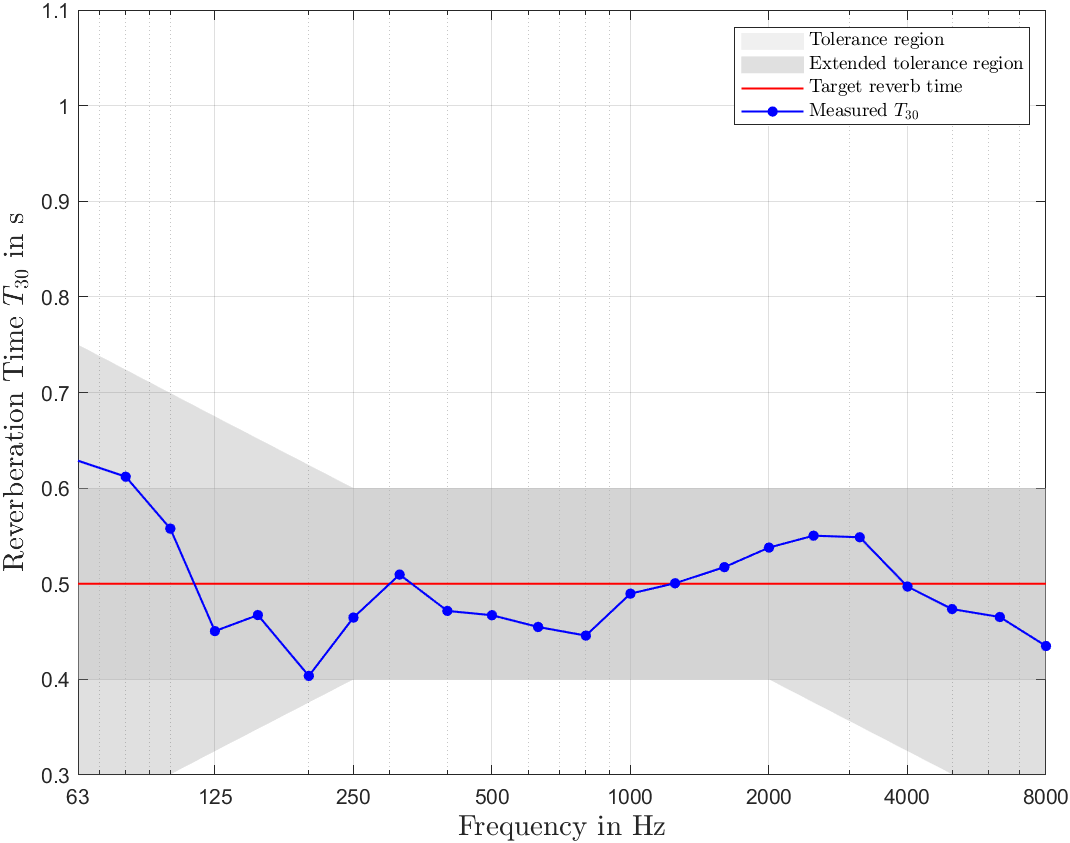}
       \caption{S2 (Blackboard): After room acoustic intervention.}
       \label{fig:T30_d}
   \end{subfigure}
   
   \centering
   \begin{subfigure}[b]{0.45\textwidth}
       \includegraphics[width=\textwidth]{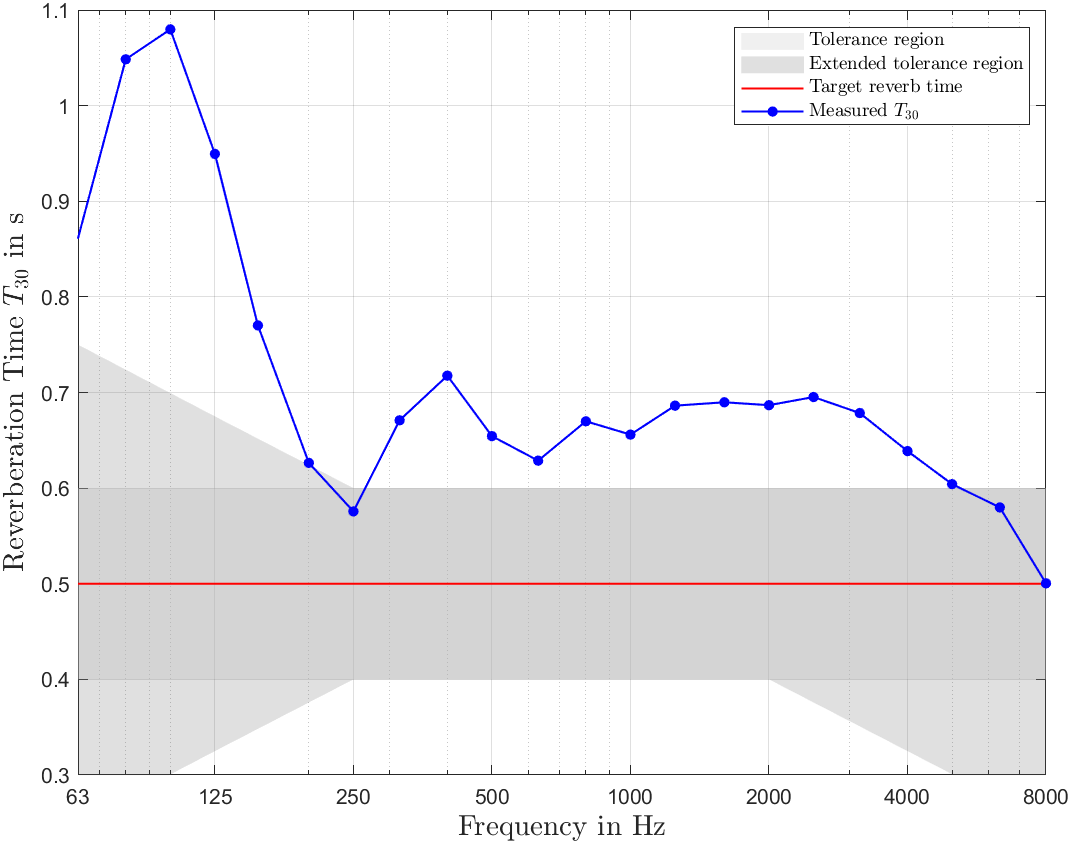}
       \caption{S3 (Center): Before room acoustic intervention.}
       \label{fig:T30_e}
   \end{subfigure}
   \hfill
   \begin{subfigure}[b]{0.45\textwidth}
       \includegraphics[width=\textwidth]{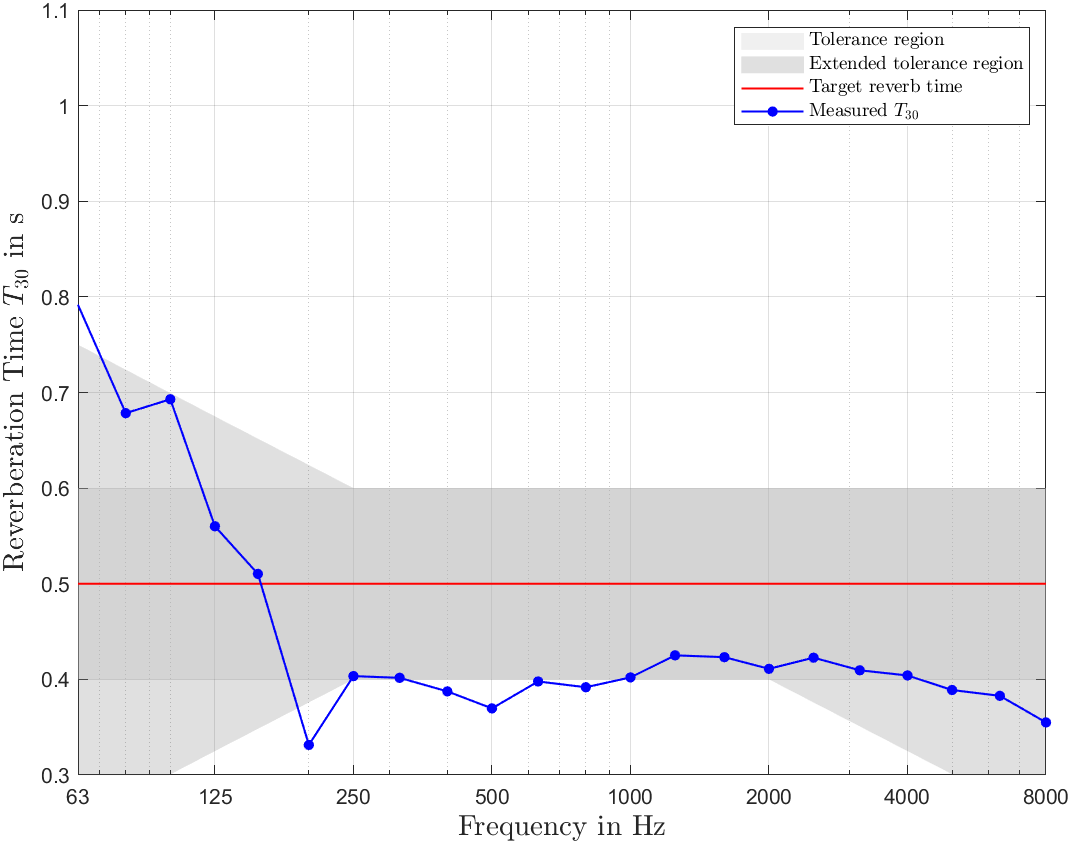}
       \caption{S3 (Center): After room acoustic intervention.}
       \label{fig:T30_f}
   \end{subfigure}
    \caption{Comparison of spatially averaged $T_{30}$ values before and after the acoustic intervention for three sound source positions (Webex Board, Blackboard, and Center). The horizontal red line indicates the target reverberation time, while the shaded gray area represents the tolerance region specified in \textcite{oe_8115}.}
    \label{fig:T30}
\end{figure*}

Figure \ref{fig:T30} compares our spatially averaged $T_{30}$ results before and after room acoustic intervention of our three measurement scenarios. The horizontal red line indicates the target reverberation time of $0.5\,\text{s}$, while the shaded gray area represents the tolerance region specified in \textcite{oe_8115}. Viewing our $T_{30}$ results before room acoustic intervention, we notice our values to be mostly above the recommended region with peaks up to $1 - 1.1\,\text{s}$ at around $100\,\text{Hz}$. This peak is consistently present throughout all measurement positions. $T_{30}$ values at middle frequencies range from $0.6 - 0.7\,\text{s}$. With higher frequencies they decrease towards the target region. This finding suggests that reverberation time in the room is too high and therefore not suitable for speech communication. Following the room acoustic intervention, we observe a reduction in reverberation times across the frequency spectrum, with most $T_{30}$ values falling within the target range. In the S2 scenario with the sound source positioned at the blackboard (Figure \ref{fig:T30_d}), $T_{30}$ values lie completely within the target region. The S1 and S3 scenarios however (Figures \ref{fig:T30_b} and \ref{fig:T30_f}) reach slightly below the target at middle frequencies. Because these two measurements were not entirely conducted in accordance to the standard, the target should be treated as a guideline instead of a strict requirement. Overall, we consider the $T_{30}$ results as satisfactory following the acoustic intervention.

\subsubsection{Speech Clarity Index $C_{50}$}
\cite{sti} states that a $C_{50}$ value of $\geq 0\,\text{dB}$ is generally presumed to indicate good speech intelligibility. However, recommended optimal $C_{50}$ values vary depending on the room's purpose and size. For example, the Italian standard UNI 11532--2 suggests an optimal $C_{50}$, $0.5 - 2\,\text{kHz} \geq 2\,\text{dB}$ for unoccupied rooms used for speech with volumes less than $250\,\text{m}^{3}$ \parencite{Improved}. Therefore, we set our initial $C_{50}$ target to above $0\,\text{dB}$ with an optimal goal of $C_{50} \geq 2\,\text{dB}$.
\par
\begin{figure*}[t]
   \centering
   \begin{subfigure}[b]{0.48\textwidth}
       \includegraphics[width=\textwidth]{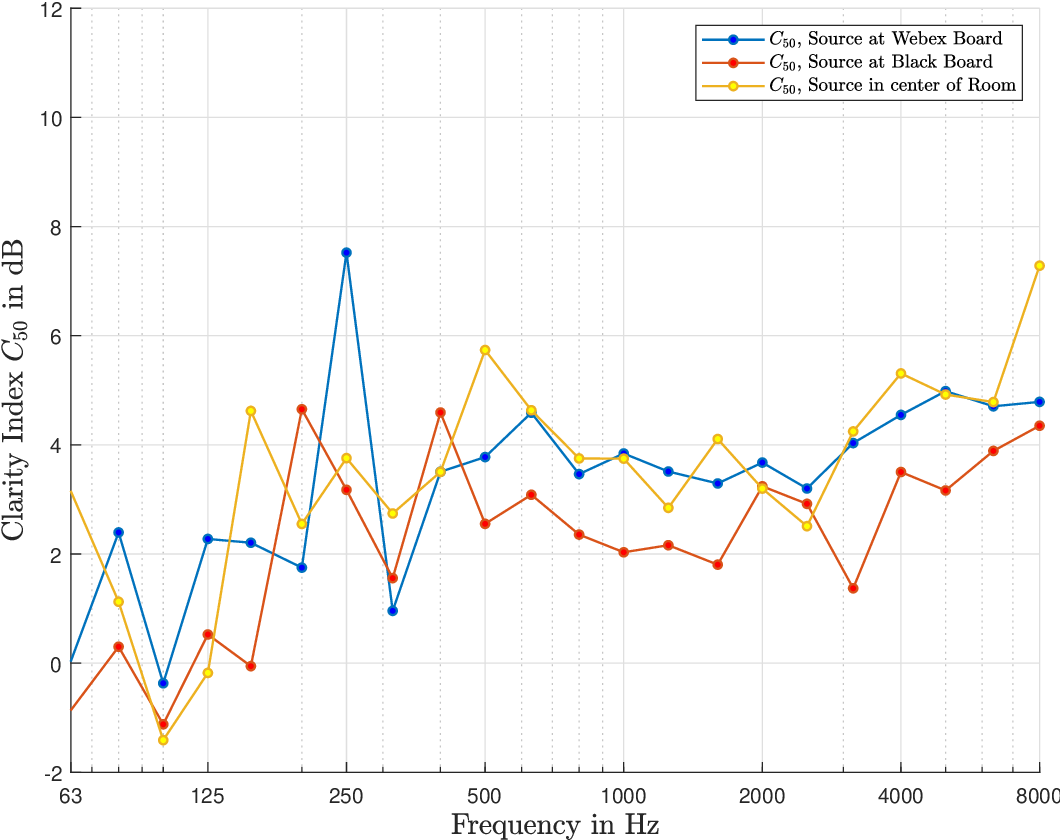}
       \caption{Spatial averaged $C_{50}$ values before acoustic intervention.}
       \label{fig:subfig1}
   \end{subfigure}
   \hfill
   \begin{subfigure}[b]{0.48\textwidth}
       \includegraphics[width=\textwidth]{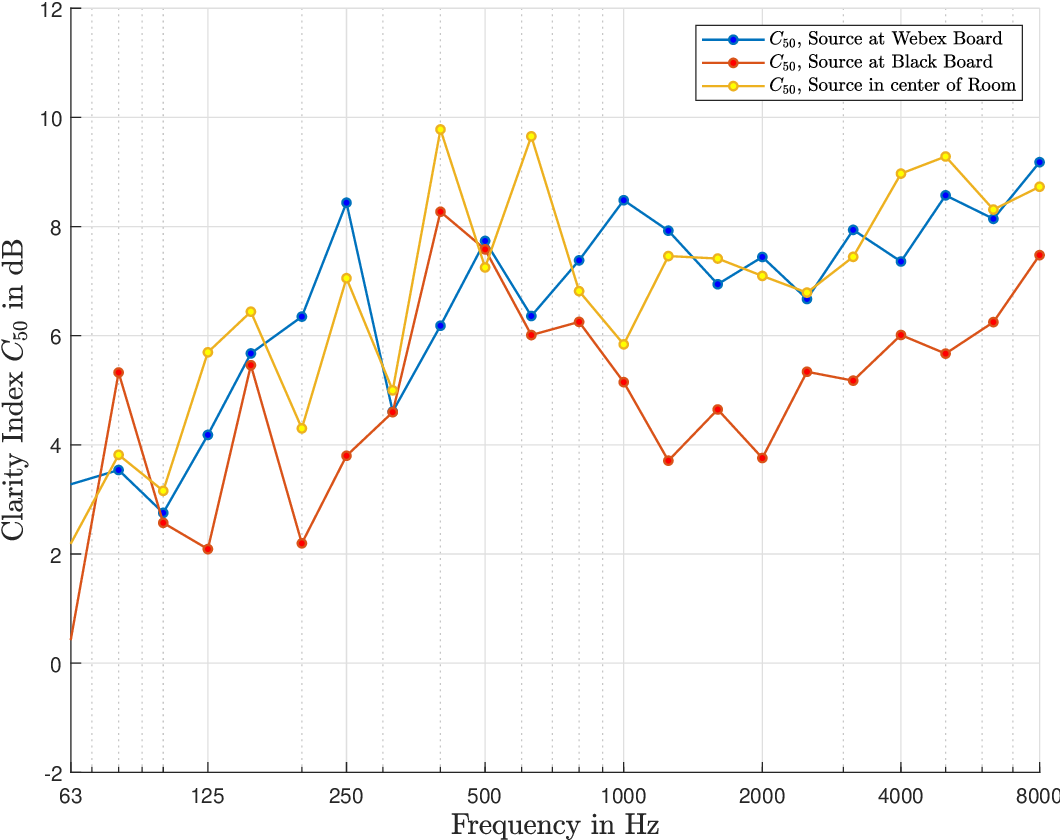}
       \caption{Spatial averaged $C_{50}$ values after acoustic intervention.}
       \label{fig:subfig2}
   \end{subfigure}
   \caption{Comparison of spatially averaged clarity index $C_{50}$ values before (a) and after (b) the room acoustic intervention of the three measurement setups.}
   \label{fig:C50}
\end{figure*}

Figure \ref{fig:C50} compares our spatially averaged $C_{50}$ results before and after acoustic intervention. The three different line colors represent the different source positions in the seminar room. Before the intervention, $C_{50}$ values were mostly above $2\,\text{dB}$, except at frequencies below $125\,\text{Hz}$, where they fell slightly below $0\,\text{dB}$. After the room acoustic intervention, $C_{50}$ values have improved across the frequency spectrum, exceeding $2\,\text{dB}$ at all frequencies. Above $250\,\text{Hz}$ they even surpass $4\,\text{dB}$, with measurements at the Webex Board and the blackboard reaching over $6\,\text{dB}$ at $400\,\text{Hz}$. Overall, $C_{50}$ values were acceptable prior to the intervention, and the room acoustic treatment enhanced speech clarity to a satisfactory level. 

\subsubsection{Speech Transmission Index $\STI$}
In contrast to $T_{30}$ and $C_{50}$, the $\STI$ considers both reverberation time and the $\SNR$. Evaluating $\STI$ results requires careful consideration of the context. Generally, for lecture rooms, an $\STI$ of $\geq 0.60$ is desirable, ideally greater than $0.75$ for optimal listening conditions \parencite[]{oe_8115}. \cite{Improved} recommends $\STI$ values higher than $0.64$ for meeting rooms and classrooms. However, in higher education settings, which often involve complex academic language, $\STI$ values below $0.75$ can be considered insufficient for listeners' speech understanding, as they impose higher cognitive demands \parencite[]{academic}. The presence of Webex in a hybrid meeting setting further increases the demand for clear acoustics. Signal processing algorithms used in remote meetings can introduce distortions which generate mismatches. This reinforces the need for $\STI$ values greater than $0.75$ \parencite[]{elu}. Although an $STI \geq 0.75$ value is highly advised for our room, achieving this consistently across all listening positions is not realistic. This goal was also not met in another study of university classrooms \parencite[]{academic}. Considering this finding, we will still aim for an $STI \geq 0.75$ target, which is a "excellent" rating according to Table \ref{tab:STI-ratings}.
\par
The $\STI$ calculation method also introduces gender-weighting factors. In our results, however, we have not seen a substantial difference between the two. Female weighted $\STI$ values were generally only approximately $0.003$ higher than male weighted values. For that reason, we decided to only show male weighted $\STI$ results, because they are slightly lower than that of females.
\par
\begin{figure*}[t]
  \centering
  \begin{subfigure}{\textwidth}
    \centering
    \includegraphics[width=\textwidth, height=0.45\textheight, keepaspectratio, trim=120 0 120 0,clip]{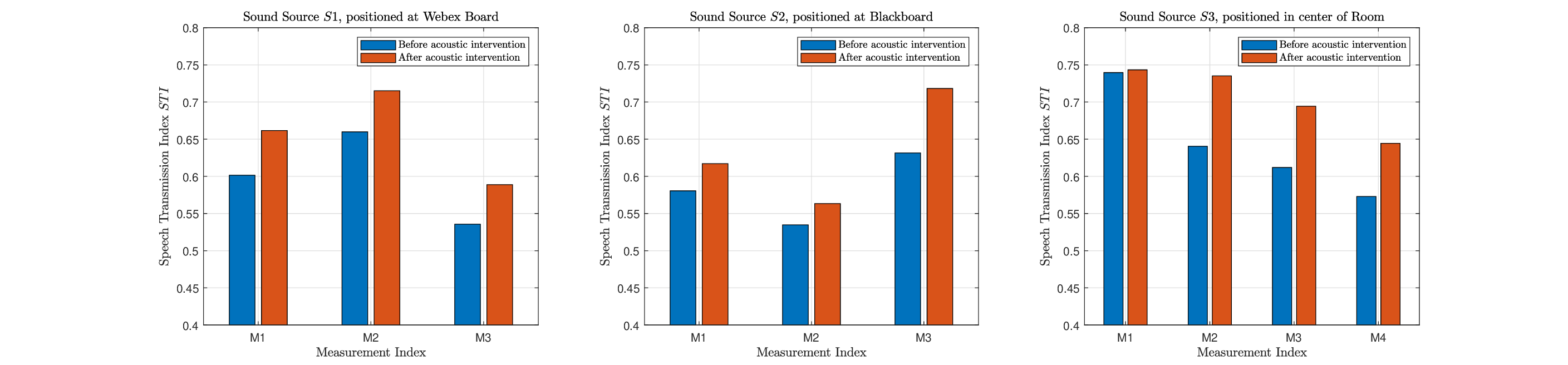}
    \caption{$\STI$ values calculated with noise generated by the projector}
    \label{fig:STI_ind}
  \end{subfigure}

  \begin{subfigure}{\textwidth}
    \centering
    \includegraphics[width=\textwidth, height=0.45\textheight, keepaspectratio, trim=130 0 130 0,clip]{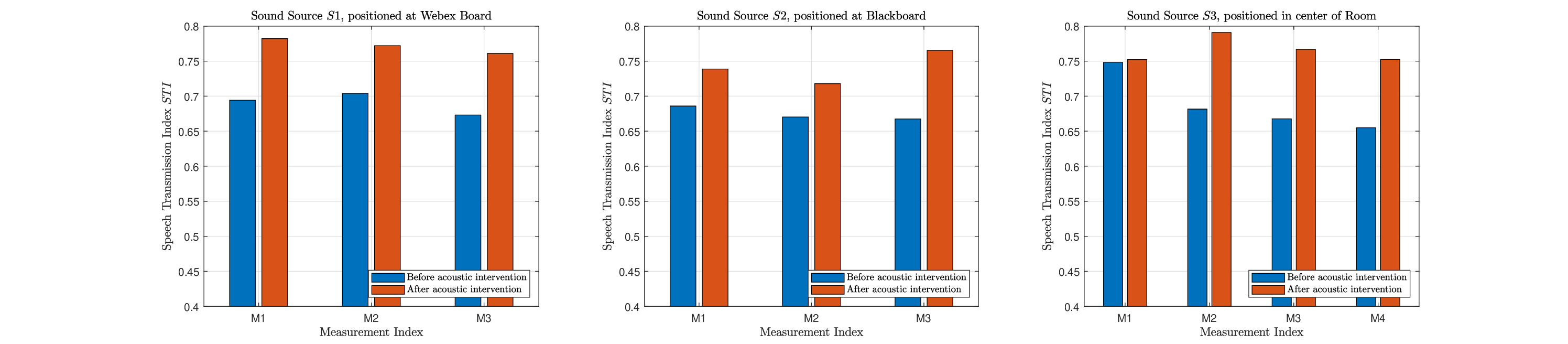}
    \caption{$\STI$ values calculated without noise generated by the projector}
    \label{fig:STI_ind_without_projector}
  \end{subfigure}

  \caption{Comparison of $\STI$ values before (blue) and after (red) the room acoustic intervention, with (a) and without (b) projector noise. Each bar graph represents one of the three measurement scenarios (S1, S2, S3) with measurement index indicating the position of the microphone in the room.}
  \label{fig:STI_combined}
\end{figure*}

Figure \ref{fig:STI_ind} shows three bar graphs which compare our individual $\STI$ results before and after room acoustic intervention, separated in our three different measurement scenarios. The measurement index describes the measurement position of the microphone, which is displayed in the schematic in Figure \ref{fig:Meas_setup}. Comparing the results before and after acoustic intervention, we observe an $\STI$ improvement of approximately $+\,0.06$ to $+\,0.09$. The largest improvement occurs for the S3 configuration with the sound source placed at the center of the room. However, none of the $\STI$ results exceed the target value of $0.75$. Even at M1, located directly next to S3, the maximum $\STI$ reaches only $0.74$, falling just short of the target. The measured positions with the lowest $\STI$ values are M3 in S1 configuration and M2 in S2 configuration. These are the most disadvantageous seating positions in terms of speech intelligibility according to our results. Both of the positions only receive ``fair" ratings, which is probably caused by them being the furthest away from their sound source. Because we suspected that the low $\STI$ ratings were strongly influenced by the noise generated by the projector, we aimed to further analyze the extent of its impact.
\par
To assess the impact of the noise generated by the projector, we analyzed $\STI$ values excluding its contribution from the calculation. An increase in background noise will directly lead to a reduction in the $\STI$, indicating poorer speech intelligibility. The $\STI$ is strongly influenced by the $\SNR$ \parencite{oe_8115}. A loud projector increases the noise floor, thereby decreasing the effective $\SNR$. This additional noise makes it particularly challenging to meet the already high demands of $\STI$ levels $\geq 0.75$. Figure \ref{fig:STI_ind_without_projector} shows these calculated values. By comparing the results of Figure \ref{fig:STI_ind} with Figure \ref{fig:STI_ind_without_projector}, we noticed massive improvements. Now all $\STI$ values after the room acoustic intervention reach above $0.70$, with most even reaching the target value $\geq 0.75$. Only M1 and M2 of the S2 configuration are slightly below the target. This finding confirms that the noisy projector heavily impairs our $\STI$ values and by eliminating the noise source, we are mostly able to reach the $\STI$ target value.
\clearpage
\newpage

\clearpage
\newpage

\section{Communicative Success in Task Oriented Dialogues in a Hybrid Space}

We measured various room acoustic metrics in order to obtain an objective assessment of the seminar room, the ultimate goal is to create an environment that not only meets technical standards but is also positively perceived by its users. To investigate the subjective impact of the acoustic intervention on speech communication quality, we invited participants to perform speech tasks in both acoustically treated and untreated versions of the seminar room. These tasks were designed to elicit natural conversation between participants. We recorded their speech\footnote{\textbf{Ethical considerations:} The speech and video were recorded after explicit written consent was obtained from the participants to use the recordings in anonymized form for research purposes. We adhered to GDPR guidelines and the European Code of Conduct for Research Integrity.} and subsequently administered a questionnaire assessing perceived communication quality. Although the scope of the experiments is limited, they provide initial evidence regarding whether the acoustic modifications improve perceived room quality.

\subsection{Experimental Setup}
All experiment runs were conducted in the same room, with participants divided into two groups of three. One group consisted of sound engineering students, who generally have a more trained ear for room acoustics, while the other group comprised individuals with no specific knowledge of acoustics. It was also important that both groups included males and females, due to differences in the frequency characteristics of the voice. Both groups were kept unaware of the actual goal of the experiment to simulate a more realistic scenario and obtain more reliable results.
\par
One participant from each group was placed in a separate room equipped with a Webex setup, including a microphone, headset and stable internet connection, and connected remotely to the seminar room. The other two participants were seated two rows in front of the Webex screen, in the center of the room. Figure \ref{fig:Experiment_SeatingPositions} shows the seating arrangement in the seminar room. The meeting was screen-recorded via Webex. Additionally, the participants’ voices were recorded individually using AKG HC 577 L microphones, while the remote participant was recorded through their headset, all in Audacity. To simulate a standard meeting in this room, the projector was turned on during the sessions, producing constant background noise.
\par
During the experiment, participants were required to solve collaborative tasks that involved as much speaking as possible. The entire experiment was documented using audio and video recordings. Each meeting lasted approximately 40 minutes and proceeded uninterrupted. Immediately after the experiment, each participant completed an individually designed questionnaire, independently and without discussing it with others or with the researchers.
\par

\begin{minipage}[t]{0.45\textwidth}
  \centering
  \includegraphics[width=0.90\linewidth, trim=100 0 100 0,clip]{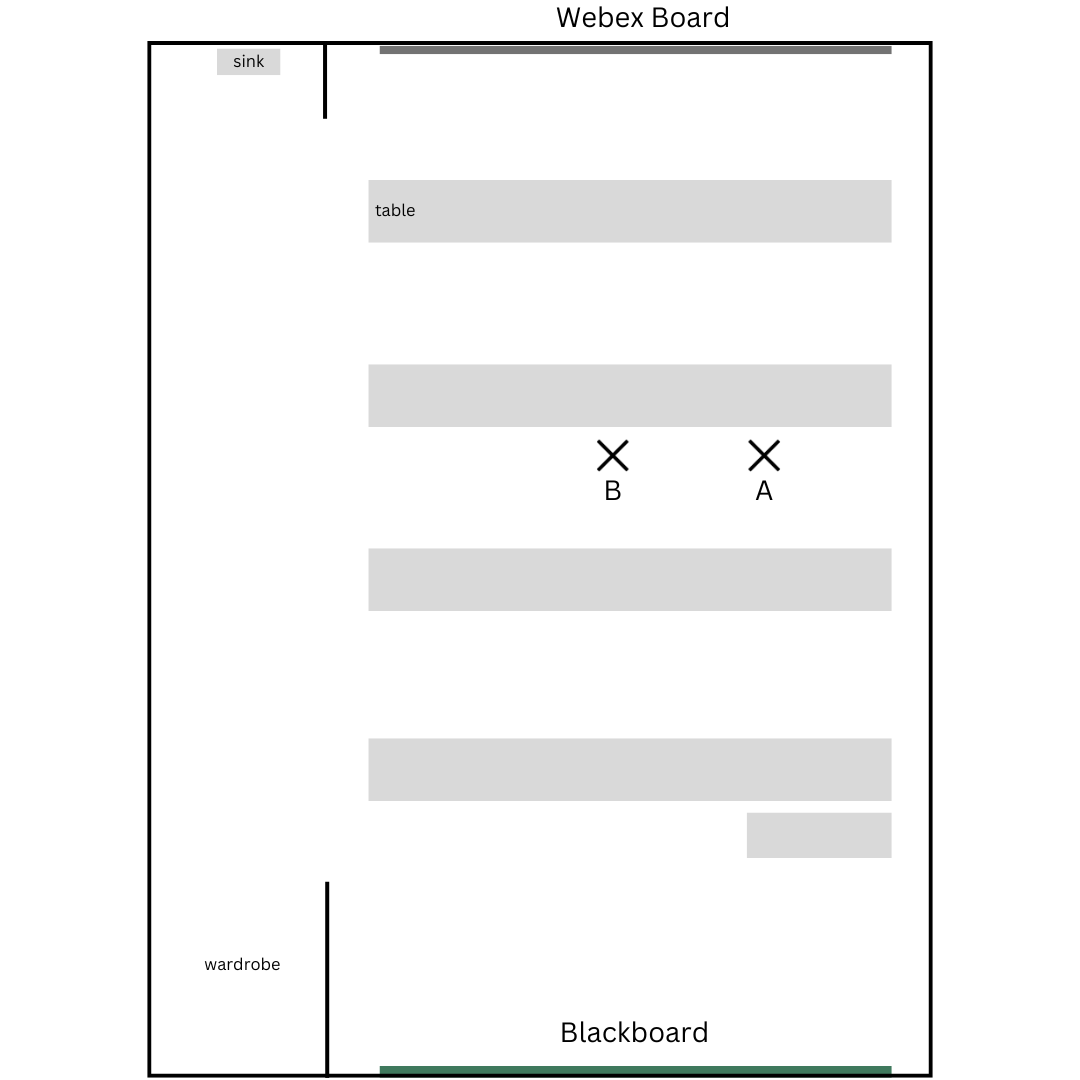}
  \captionof{figure}{Seating positions of participants A and B in the seminar room during the experiments. The participants are facing the webex screen.}
  \label{fig:Experiment_SeatingPositions}
\end{minipage}

\paragraph{Experiment 1: Before room acoustic intervention}
The first experiment was conducted before any acoustic improvements were applied to the room. This allowed us to ask participants about their experience in the untreated room.

\paragraph{Experiment 2: After room acoustic intervention}
The second experiment was conducted in the same room as the first, but after the acoustic improvements had been implemented. The groups performed similar tasks and completed the questionnaire again, rating their experience of the treated room and any perceived differences. Participants were not informed about the room treatment or the specific purpose of their participation (although for the sound engineering students, the visual presence of acoustic modifications may have made the intervention apparent).

\subsubsection{Participants}
Two participant groups were formed: one with acoustically inexperienced individuals (G1) and one with acoustically experienced individuals (G2). Within each group, participants were assigned an identifier (A, B, or C), with C denoting the remote participant. This resulted in six identifiers: G1A, G1B, G1C, G2A, G2B, and G2C.

Due to time constraints and the scope of the project, participants were recruited from among colleagues and friends. In total, six students participated, aged between 20 and 30 years. Three were female and three were male. All participants were native German speakers to avoid language barriers that could interfere with the experiment.

\subsubsection{Communication Tasks}
To simulate a realistic meeting situation, we prepared two tasks for participants to complete during each session. Within the limited time, participants were encouraged to talk as much as possible with one another. The tasks were designed as engaging games that required active conversation between participants. Two types of games were selected to generate varied and dynamic interactions.

\paragraph{Game 01: Taboo}
The first game follows the rules of the popular game Taboo. One player receives a word to explain to the other players using paraphrasing. In our version, each player was given twelve words to describe, taking turns after each word. This maintains a balanced turn-taking dynamic, although differences in individual knowledge still creates some imbalance. The game’s core mechanism relies heavily on conversation, making it ideal as the first task.

\paragraph{Game 02: Connection}
The second game, Connection, requires participants to work as a team to identify common threads between words. Players receive a grid of sixteen words and are tasked with dividing them into four groups of four words that share a common theme. Each group have four puzzles to complete in total, with increasing difficulty. Unlike Game 01, all participants have to collaborate as a team, producing a different interaction dynamic and greater variety in conversational forms throughout the experiment.
\par
Each meeting session lasted between 30 and 40 minutes. Both games were designed to take approximately 15 minutes each. The rules were printed and provided to the remote participant, who explained them via Webex. For the experiments conducted before and after the acoustic intervention, the game rules remained identical, although the specific content differed. This ensured maximum comparability between the two experiments, as the only intended variable was the change in room acoustics.

\subsubsection{Evaluation via Questionnaire}
Each participant completed a questionnaire (see German original in Appendix \ref{sec:appendix}) that collected both general background information and detailed evaluations of their experience during the assigned tasks. The first section covered languages spoken, any linguistic training, upbringing, and time spent abroad, providing insights into participants’ speaking experience and vocal tract characteristics.
\par
The main survey focused on participants’ subjective experience, with most questions indirectly addressing room acoustics to avoid biasing responses. It began with two open-ended questions, asking participants to describe their experience playing the games, the effort required to cooperate with others, and which participants were more difficult to understand.
\par
Next, a series of scale questions assessed participants’ evaluations of the experiment in both the untreated and acoustically treated room. These items addressed perceived background noise, ability to concentrate, frequency of repetitions by themselves or others, and ease of working remotely.
\par
The final section asked participants to suggest improvements and report on their prior familiarity with fellow participants before the first experiment. For the second experiment, an additional open-ended question and two scale questions were included to assess whether the experience had improved and what changes participants noticed compared to the first session.

\subsection{Results}
The goal was to collect a comprehensive set of recordings, questionnaires, and other data to serve as a resource for future studies and to generate further insights. After completing both experiments, we analyzed the results and organized them into three categories for clarity. The first category summarizes the scale-based responses, facilitating easy interpretation and comparison. The second highlights key phrases and observations drawn from the questionnaire analysis. The third presents additional data collected outside of the questionnaire.

\subsubsection{Evaluation based on scale questions}
Table \ref{tab:qresults} summarizes all scale questions from the first and second experiments. The column "Average Rating E1" shows the averaged values of both groups for the first experiment, while "Average Rating E2" presents the averages for the second experiment. Question 7 is excluded, as it cannot be represented as an averaged result. The scale questions range from 1 (not true at all) to 8 (completely true), with the average score listed next to each question. Ratings for Q2, Q3, Q5, Q6, and Q9 were inverted so that higher scores consistently reflect better quality; the table presents these questions in their translated form.
\par
\vspace{+15pt}
\begin{minipage}{\columnwidth}
    \captionsetup{width=0.9\columnwidth} 
    \captionof{table}{Summary of averaged questionnaire ratings for the first and second experiments. Values represent the combined responses of all participants across both groups. The original items were presented in German and translated for reporting}
    \label{tab:qresults}
    \resizebox{0.95\columnwidth}{!}{%
        \begin{tabular}{|p{0.6\columnwidth}|p{0.2\columnwidth}|p{0.2\columnwidth}|}\hline
             \textbf{Question} & \textbf{Average Rating E1} & \textbf{Average Rating E2} \\ \hline
             Q1: I found it easy to complete the task with my group. & 6.0 & 7.2\\ \hline
             Q2: I did not perceive the background noises as unpleasant or disturbing. & 7.3 & 7.7\\ \hline
             Q3: I did not perceive the overall noise environment as unpleasant or disturbing. & 7.0 & 6.8\\ \hline
             Q4: I was able to concentrate at any moment of the experiment. & 6.7 & 7.5\\ \hline
             Q5: I did not feel the need to repeat myself in order to be understood. & 7.2 & 4.3\\ \hline
             Q6: I did not perceive that others were frequently misunderstood. & 6.8 & 6.0\\ \hline
             Q8: Working remotely with somebody was easy. & 6.5 & 7.2\\ \hline
             Q9: I was able to understand the remote person better than the person inside the room.\footnotemark[3] & 5.0 & 4.0\\ \hline
             Q10: There was a noticeable improvement in this experiment regarding both the noise environment and understandability. & n/a \footnotemark[4]  & 5.2 \\ \hline
             Q11: I was able to concentrate much better in this round of the experiment, particularly when solving the task. & n/a \footnotemark[4]  & 5.3\\ \hline
        \end{tabular} 
    }
\end{minipage}
\vspace{+10pt}

The questionnaire results indicate an overall improvement in participants' experience following the room acoustic intervention. Most questions measuring task ease, concentration, and perceived noise (Q1, Q2, Q4, Q8) showed higher average ratings in the second experiment, suggesting that the acoustic modifications positively affected participants’ ability to focus and communicate. Particularly notable is Q8 (``Working remotely with someone was easy"), which increased by nearly one point, and Q4, which also showed a marked improvement. The newly added questions in the second experiment (Q10, Q11) received ratings above 5, indicating that participants perceived improvements in noise environment, understandability, and concentration.

\footnotetext[3]{Only answered by participants inside the seminar room.}
\footnotetext[4]{Question was not in Experiment 1.}
\par
\begin{figure*}[t]
    \includegraphics[width=\linewidth]{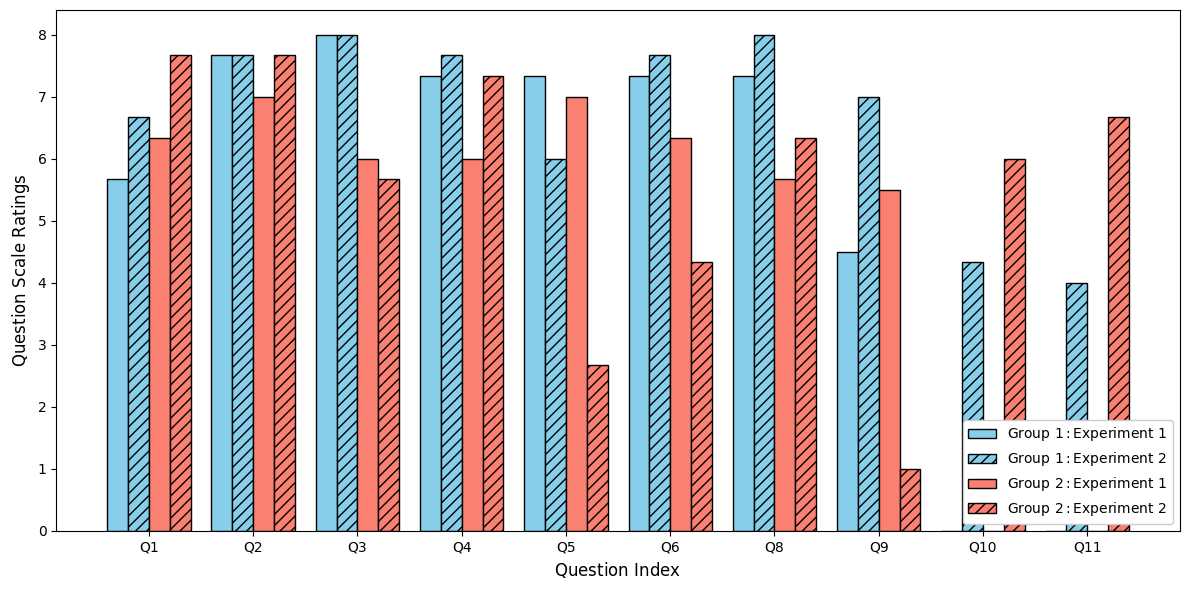}
    \caption{Comparison of the averaged questionnaire scale ratings of Group 1 (blue) and Group 2 (red) between the two experiments. Experiment 1: before room acoustic intervention, Experiment 2: after room acoustic intervention.}
    \label{fig:survey_bar}
\end{figure*}

We also analyzed the differences in both groups' answers. Figure \ref{fig:survey_bar} shows the averaged results of both groups from the questionnaire. 

Conversely, some questions related to self- and peer-understanding (Q5, Q6, Q9) decreased after the intervention. For example, Q5, which asks whether participants had to repeat themselves, was rated lower in the second experiment. G2, the acoustically experienced group, rated the second experiment as less disturbing and reported higher ease of concentration, while Q9 showed a decrease for this group in the first experiment compared to the second. Overall, the results suggest that the room acoustic intervention generally enhanced participants' experience and performance in hybrid meeting tasks, while highlighting variability across specific aspects of perceived communication quality.

\subsubsection{Evaluation based on verbal feedback}
Next, we evaluate those aspects of the questionnaire that participants responded in their own words. This delivers further understanding than the scale questions. In the following, we list those answers that stood out as they were mentioned by more than one participant in a similar form: 

\begin{enumerate}
    \item Speaking at the same time as counterproductive and problematic. This is mentioned several times in response to our question "How hard was it to cooperate with the other participants?" after the first experiment.
    \item Face-to-face communication seems to be the preferred option since it is highlighted multiple times, particularly in the first experiment. Since participants were unaware of the experiment's purpose, they questioned the need for a hybrid meeting situation. This relates to our second research question RQ2: "Does communication success in hybrid meetings improve when room acoustics are enhanced?". In contemporary work environments, remote collaboration is often unavoidable or more efficient, which largely resolves the question of why hybrid meetings are conducted. Therefore, the focus should shift to how remote interactions can be improved to resemble face-to-face communication more closely. The responses of our participants underline the importance of this issue and reflect the expectations associated with remote work.
    \item  Another topic is the projector inside the room. At least one participant of every group mentioned the constant distracting background noise produced by it. This confirms the results of the $\STI$ in section 2 and highlights the importance of controlling background noise in a hybrid meeting space. Any kind of device operating at a certain volume level has the potential to disrupt a meeting and should be considered when setting up a hybrid meeting space.
\end{enumerate}
\par
We now cite selected participant responses to support our observations and results. Participant G1C, who joined the meeting remotely, states: ``In general, it wasn't tiring or complicated. I understood both participants well, except if they spoke at the same time." And Person G1B inside the room mentioned: ``It was not really tiring, though if I have to point out someone, it is the remote person, just because it does not feel so natural." Both responses are translated, as the participants’ primary language is German. These responses reinforce points (1) and (2) of our earlier discussion. We observe similar patterns in Group 2 (G2). 
\par
We again cite G1B from the acoustically inexperienced group, who describes the second game as ``frustrating" in the first experiment (untreated room). When asked if it was tiring to cooperate in the second experiment (acoustically treated room), the participant answered: ``It was not tiring to cooperate with the others. It was even fun. No noticeable difficulties." And then also for the second experiment the same person (G1B) mentioned: ``The last time it was more disturbing with one person remote. This time, also regarding the audio quality, it wasn't even noticeable. For solving the games, this time, the second game was much more pleasant."
\par
This illustrates a significant observation: Person G1B was unhappy with the second game in the first experiment and identified the remote participant as most difficult to understand. In the second experiment, these issues do not appear. Similar patterns appeared in the answers of other participants. The most informative responses were obtained from questions addressing the comparison between the second experiment with the first. We conclude with the words of participant G2C: ``Background noises were reduced. Concentration was focused on the voice, which was beneficial."

\paragraph{Other Observations}
Naturally, not all aspects proceeded as anticipated, and the same applies to our survey and experimental setup. Some minor issues were observed after analyzing the questionnaire responses and through discussions with participants following the second experiment.

The first issue concerns the projector, previously identified as a significant noise source. To simulate realistic conditions, the projector was switched on during the experiments; however, it occasionally entered energy-saving mode and switched off. Some participants noted that the absence of projector noise improved their experience. It is important to clarify that the projector is not an integral part of the hybrid meeting setup in this room. It serves as an additional option for presentations and projections onto the side of the room opposite the meeting screen.

A second issue relates to the audio technology used in both experiments. Consistent audio levels were not maintained throughout the meetings, including the remote participant's headset, and different headset models were employed.

\section{Discussion}

\paragraph{Room Acoustic Measurements} As the results of the two measurements show, a clear improvement was achieved in the treated room. Since this is a university-owned room, for administrative reasons, we could not implement permanent changes. The materials available for this study were limited to existing university resources, usable only on a temporary basis during the experiment. This fact limits our advances, as ceiling-mounted elements would also have been be an important part of the room acoustic treatment. Despite these limitations, we showed a clear improvement, illustrating that it would be easily possible to find also a permanent solution. 
\par
One of the bigger challenges in this particular room was the limited amount of available space. The architectural and interior design is relatively complex, which complicates the optimal placement of acoustical panels. In future work it will be essential for the design of permanent interventions to also employ acoustical 3D software for simulating room acoustics and optimizing treatment placement. The best positions for bass traps in the corners are rather simple to find, due to the room's restrictive layout. However, acoustic software is key to find optimal placement for wall and ceiling absorbers and diffusers. Additionally, it would be beneficial to identify the most efficient approach, one that minimizes material, while still providing a high quality meeting environment. Although less critical, the visual appearance remains an open important issue, since the acoustic treatment should integrate seamlessly without drawing too much attention.

\paragraph{Task-oriented dialogues in hybrid meetings}

In this study we obtained high-quality recordings of all experimental meetings (audio and video, at site and via the used webex platform), which were not yet analyzed for the current work. This represents a significant opportunity for future research. At SPSC laboratory, we plan to annotate and analyze these recordings, allowing us to gain detailed insights into participant interactions, speech intelligibility, and Lombard speech. Our analysis will include the comparison not only between the experimental conditions but also between on-site and remote speakers, as well as across different participants and tasks to increase our understanding of acoustic treatment effects on communication quality and conversational dynamics in hybrid meetings.

Despite the small number of participants in our study, the observed tendencies in the questionnaire clearly show a benefit of the room acoustic treatment but also that in both conditions, the remotely participating person experiences communication less effective than the persons being at site in the seminar room. In future work, we plan to record a larger number of teams with different numbers of persons per meeting, as we expect different effects on communication quality depending on the number of persons on site and participating a meeting remotely. Furthermore, it would be important to ensure a broader demographic coverage including variations in age, gender, educational and professional background. Secondly, also the duration and the content of the experiment will have to be more naturalistic in future work. We plan to record persons in their weekly work meetings over a longer period of time (one year), starting out in the acoustically untreated room, and then changing to the treated room. These planned large-scale experiments will indicate not only the actual impact of room acoustic interventions but also support the identification of those conditions where communication needs to be supported by speech processing tools.

\section{Conclusion}
This study investigated for a specific seminar room at TU Graz, whether it met the requirements for effective use in a
hybrid meeting and whether room acoustic interventions can enhance communication quality in hybrid meetings. The initial measurements revealed several acoustic deficiencies in the room under study. Both $T_{30}$ and $\STI$ values fell short of recommended standards. Reverberation times were consistently above target levels across all frequency bands and measurement conditions, with particularly pronounced deviations below $125\,\text{Hz}$. $\STI$ values ranged between “fair” and “good,” depending on listening position, whereas an “excellent” rating would be desirable given the elevated cognitive demands associated with the room’s intended use. 
Our room acoustic measurements showed that the room did not meet the requirements for effective use in a hybrid meeting, which was in line with the answers of the participants in the questionnaires, indicating also perceived acoustic problems.
\par
Equally important is the evaluation of how participant assessments change after the acoustic intervention. Even if the experiments do not reveal a clear quantitative improvement, noticeable trends were observed. Overall, participants reported a more positive experience, reflecting the beneficial impact of the intervention.
The objective room-acoustic measurement results, however, demonstrate a clear improvement compared to the untreated room. Both the clarity index and, in particular, the reverberation time now meet the target values. And the $\STI$ increases from a "fair" rating to a nearly "excellent" (without the projectors' background noise, the measures reached excellent values). Based on these results, we conclude that objective requirements for such a meeting room are improvable without excessive effort. This is true, however, under the basic assumption that we drew that the room acoustic requirements to a meeting room are the same in a hybrid meeting. An assumption given the absence of already existing standards for hybrid communication spaces. In future, we plan to conduct large-scale studies which may contribute to new standards for the room design of hybrid meeting spaces.  

\paragraph{Contributions}

Conceptualization: M.H. and B.S.; Methodology: J.K, M.H. and B.S.; Software: S.J. and J.K.; Validation: B.S. and M.H.; 
Formal analysis: R.E. and S.J. and J.S.; Data curation: R.E. and S.J. and J.S.; Writing---original draft preparation: R.E. and S.J. and J.S.; Writing---review and editing: all authors; 
Visualization: R.E. and S.J. and J.S.; Supervision: J.K., M.H. and B.S.; All authors have read and agreed to the published version of the manuscript.

\paragraph{Declaration of Generative AI and AI-assisted technologies in the writing process}
We confirm that AI-based tools (ChatGPT and NotebookLM) were used only for purposes of language improvement, style, and grammar, and not for the creation of original content or data analysis. After using this tool/service, all authors reviewed and edited the content as needed and thus take full responsibility for
the content of the publication.

\printbibliography

\section{Appendix}
\label{sec:appendix}










\myheading{1. Questionnaire}



\section*{Supplementary material}

# Experiment Umfrage

## Einleitung

Zuerst möchten wir uns bedanken, dass Sie an unserem Experiment für unsere Bachelorarbeit teilnehmen.
Jetzt brauchen wir nur noch paar persönliche Einschätzungen, deswegen ein kleiner Fragebogen. Bitte beantworten Sie die Fragen für sich selbst und rückblickend auf das Experiment, so gut wie möglich.
Alle Daten sind vertraulich und werden nur für wissenschaftliche Zwecke verwendet.

## Persönliche Informationen

- Sprecher-ID: _____

- Geschlecht: _____

- Beruf/Studium: _____

- Muttersprache: _____

- Wie viele Fremdsprachen und auf welchem Niveau:
  _____

- Haben Sie eine sprachwissenschaftliche Ausbildung (oder Gesangserfahrung)?:
  _____

- Aktueller Wohnort: _____

- Wohnort zur Volksschulzeit: _____

- Dauer der schulischen Ausbildung und Abschluss:
  _____

- Im Ausland gelebt? Wie lange und wo?:
  _____

# Erfahrungsfragen

Die folgenden beiden Fragen können Sie nun auch im Vergleich zum letzten Experiment setzen. Oder neu formulieren und einfach nur ihre jetzige Erfahrung teilen.

- Wie war Ihre Erfahrung beim Spielen der Spiele oder lösen der Aufgabe? Wie haben Sie sich gefühlt?

  _____

  _____

  _____

- Wie anstrengend war es mit den anderen Versuchsteilnehmer*innen zu kooperieren? Bei welcher Person war es am kompliziertesten und warum?

  _____

  _____

  _____

# Skalenfragen

Bewerten Sie die folgenden Aussagen und kreuzen Sie das für Sie zutreffende an
(1 = trifft überhaupt nicht zu, 8 = trifft voll zu):

1. Ich konnte die Aufgabe leicht mit meiner Gruppe lösen:

   | **1** | **2** | **3** | **4** | **5** | **6** | **7** | **8** |
   |---|---|---|---|---|---|---|---|
   | ☐ | ☐ | ☐ | ☐ | ☐ | ☐ | ☐ | ☐ |

2. Ich empfand die Hintergrundgeräusche als unangenehm und störend:

   | **1** | **2** | **3** | **4** | **5** | **6** | **7** | **8** |
   |---|---|---|---|---|---|---|---|
   | ☐ | ☐ | ☐ | ☐ | ☐ | ☐ | ☐ | ☐ |

3. Ich empfand die generelle Geräuschkulisse als unangenehm und störend:

   | **1** | **2** | **3** | **4** | **5** | **6** | **7** | **8** |
   |---|---|---|---|---|---|---|---|
   | ☐ | ☐ | ☐ | ☐ | ☐ | ☐ | ☐ | ☐ |

4. Ich konnte mich zu jedem Zeitpunkt des Experiments konzentrieren:

   | **1** | **2** | **3** | **4** | **5** | **6** | **7** | **8** |
   |---|---|---|---|---|---|---|---|
   | ☐ | ☐ | ☐ | ☐ | ☐ | ☐ | ☐ | ☐ |

5. Ich hatte das Gefühl, mich häufiger wiederholen zu müssen, um verstanden zu werden:

   | **1** | **2** | **3** | **4** | **5** | **6** | **7** | **8** |
   |---|---|---|---|---|---|---|---|
   | ☐ | ☐ | ☐ | ☐ | ☐ | ☐ | ☐ | ☐ |

6. Ich hatte das Gefühl, dass jemand anderes häufig nicht verstanden wurde:

   **1**  **2**  **3**  **4**  **5**  **6**  **7**  **8**
   ☐    ☐    ☐    ☐    ☐    ☐    ☐    ☐

7. Falls ja:

   | Wen hast du schlecht verstanden? | |
   |---|---|
   | 1. Person: _____ | ☐ |
   | 2. Person: _____ | ☐ |

8. Mit jemandem remote zu arbeiten war einfach:

   **1**  **2**  **3**  **4**  **5**  **6**  **7**  **8**
   ☐    ☐    ☐    ☐    ☐    ☐    ☐    ☐

9. Ich konnte die Personen im Raum besser verstehen als die remote Person (nur von A und B zu beantworten):

   **1**  **2**  **3**  **4**  **5**  **6**  **7**  **8**
   ☐    ☐    ☐    ☐    ☐    ☐    ☐    ☐

10. Es gab eine bemerkliche Verbesserung in diesem Durchlauf des Experiments, sowohl in Bezug auf die Geräuschkulisse, als auch die Verständlichkeit.:

    **1**  **2**  **3**  **4**  **5**  **6**  **7**  **8**
    ☐    ☐    ☐    ☐    ☐    ☐    ☐    ☐

11. Ich konnte mich in diesem Durchlauf des Experiments viel besser auf meine Mitprobant*innen und das lösen der Tasks/Spiele konzentrieren.:

    **1**  **2**  **3**  **4**  **5**  **6**  **7**  **8**
    ☐    ☐    ☐    ☐    ☐    ☐    ☐    ☐

## Vergeleich zum ersten Experiment

Was hat sich für Sie in diesem Durchlauf des Experiments geändert? Gab es irgendwelche auffälligen Verbesserungen um die Aufgaben/Spiele zu lösen?

_____

_____

_____

_____

_____

## Abschließende Frage

Was würden Sie dieses Mal ändern, um eine noch bessere Erfahrung zu haben?

_____

_____

_____

_____

_____

**Vielen Dank für Ihre Teilnahme!**





\end{document}